\documentclass[11pt,conference]{IEEEtran}

\usepackage{cite}               
\usepackage{graphicx}           
\usepackage{amsmath,amssymb}    
\usepackage{algorithm}
\usepackage{algpseudocode}      
\usepackage{booktabs}           
\usepackage{multirow}           
\usepackage{array}              
\usepackage{caption}
\usepackage{subcaption}         
\usepackage{url}                
\usepackage{hyperref}           
\IEEEoverridecommandlockouts
\hypersetup{
	colorlinks=true,
	linkcolor=blue,
	citecolor=blue,
	urlcolor=blue
}
\title{FinAgent: An Agentic AI Framework Integrating Personal Finance and Nutrition Planning%
\thanks{This paper was presented at the IEEE International Conference on Computing and Applications (ICCA 2025), Bahrain.}}

\author{
	\IEEEauthorblockN{
		Toqeer Ali Syed\textsuperscript{1,*}, Abdulaziz Alshahrani\textsuperscript{1}, 
		Ali Ullah\textsuperscript{1}, Ali Akarma\textsuperscript{1}\\
		Sohail Khan\textsuperscript{2}, Muhammad Nauman\textsuperscript{2}, Salman Jan\textsuperscript{3}
	}
	\IEEEauthorblockA{
		\textsuperscript{1}Faculty of Computer and Information System, Islamic University of Madinah, Saudi Arabia\\
		\textsuperscript{2}Department of Computer Science, Effat College of Engineering, Effat University, Saudi Arabia\\
		\textsuperscript{3}Arab Open University, Bahrain\\
		\textsuperscript{*}Corresponding author: toqeer@iu.edu.sa
	}
}

\begin{document}
	\maketitle
	
	\begin{abstract}
		The issue of limited household budgets and nutritional demands continues to be a challenge especially in the middle-income environment where food prices fluctuate. This paper introduces a price aware agentic AI system, which combines personal finance management with diet optimization. With household income and fixed expenditures, medical and well-being status, as well as real-time food costs, the system creates nutritionally sufficient meals plans at comparatively reasonable prices that automatically adjust to market changes. The framework is implemented in a modular multi-agent architecture, which has specific agents (budgeting, nutrition, price monitoring, and health personalization). These agents share the knowledge base and use the substitution graph to ensure that the nutritional quality is maintained at a minimum cost. Simulations with a representative Saudi household case study show a steady 12-18\% reduction in costs relative to a static weekly menu, nutrient adequacy of over 95\% and high performance with price changes of ±20-30\%. The findings indicate that the framework can locally combine affordability with nutritional adequacy and provide a viable avenue of capacity-building towards sustainable and fair diet planning in line with Sustainable Development Goals on Zero Hunger and Good Health.
	\end{abstract}
	

\section{Introduction}

Financial, healthcare, and digital services have also been transformed by artificial intelligence (AI) with generative AI being used to create content, reason, and support interactive decision-making \cite{brown2020fewshot, openai2023gpt4}. Nevertheless, the majority of systems are responsive and do not have long-term planning or goal-seeking independence. The gaps are filled in agentic AI, which involves combining action loops based on planning, monitoring, memory and tools, and facilitating goal-directed behaviour \cite{schneider2025generative, park2023generative}. Gartner estimates that in 2028, agentic-AI will be present in one-third of applications \cite{gartner2025agentic}.

With these developments, household decisionmaking, especially at the point of budgeting and nutritional requirements, is under-served. Current tools monitor expenses or suggest meals but do not combine financial constraints, nutritional needs and real-time prices \cite{khan2015correlating}. Low quality of diet increases long term health risks \cite{cannon2022food}, particularly among the low income families, which explains the applicability to SDG-2 and SDG-3.

This work presents a model of agentic-AI that integrates the real-time price monitoring, budgeting, nutritional maximization, and customized dietary restrictions. The system takes into account the market changes and health needs by using a multiagent architecture and cost-sensitive optimization with substitution graphs to adapt the household menus. Saudi household simulations reveal 12-18 \% of cost reduction, and more than 95 \% nutrient adequacy.


\section{Background}
\label{sec:background}

\subsection{Healthy Diet Fundamentals}

Healthy diets are not only preventive of chronic disease, but must contain balanced macronutrients, sufficient amounts of vitamins and minerals, and minimal amounts of sugars, saturated fats, and sodium \cite{who2021diet, lips2021vitamind}. FAO focuses on different kinds of foods such as fruits, vegetables, legumes, whole grains, lean proteins, and dairy \cite{fao2022nutrition}. Iron, calcium and omega-3 fatty acid deficiencies are still common across the globe \cite{allen2003micronutrient}. Cultural and religious considerations including halal rules and Ramadan fasting also influence the preferences of the diet and promote flexible and individualized planning of meals \cite{shatila2021ramadan}.

\subsection{Personal Finance for Households}

Financial management is important to the stability of the household. Whilst guidelines like the 50/30/20 rule are helpful in providing a framework, they might not be suitable to low and middle-income families \cite{yeo2024financialplanning}. In most middle-income areas, the proportion of food to income is 15-25 \% and increases with inflation \cite{worldbank2023foodsecurity}. By combining nutrition and budgeting, food resources, which are limited in number, can be allocated more effectively.

\subsection{Agentic AI Concepts}

Agents AI combine perception, memory, planning and self-monitoring in a loop \cite{bandi2025rise}. It drives the retail, travel, and shopping agents that operate round the clock, but most of them are aimed at convenience, not health or financial welfare \cite{bandi2025rise}. A household decision support is a new direction that can be achieved through integrating nutritional science, economic modeling, and dynamic pricing into an agentic architecture.

\section{Related Work}
\label{sec:related}

Literature on the relationship between household finance and optimal diets is scarce, encompassing FinTech, diet planners, and agentic AI.

\subsection{Budgeting and Financial Technologies}

FinTech apps (Mint, YNAB, PocketGuard) track revenues and sort expenditures but are reactive and hardly take nutrition or health into account \cite{suryono2020fintech}. Newer systems combine constrained health information with no dynamic adjustments of prices or supplies \cite{kayikci2022pricing, martell2024inflation, nalawade2024reinforcement}. These are further extended by our framework to optimize budgets and nutrition jointly in the face of changing prices in real-time.

\subsection{Diet Planners and Nutritional Optimization}

The classical least-cost models \cite{stigler1945cost} have developed to include nutrient databases and sustainability restrictions \cite{babalola2020diet, verly2022sustainable, rocabois2022indigoo, acharya2025price}. The majority of them disregard domestic realities, including shared budgets, multigenerational requirements, cultural inclinations, and variable prices, and do not offer personalization. These gaps are filled with our system with real-time, priceaware dietary optimization.

\subsection{Agentic AI Systems in Consumer Domains}

Agentic AI aids retail \cite{bensaad2023virtualagent}, travel \cite{yang2022multiagent}, and consumer assistance \cite{tarigan2025llmagent}, however there is little domestic finance or nutrition.Current diet applications are generic chatbots that do not have financial logic, personalization, or adaptive replanning \cite{oh2021chatbots}. Earlier-developed work is siloed: FinTech applications are used to handle budgets, nutrition applications handle intake, and agentic systems handle individual tasks. FinAgent incorporates autonomous replanning, cultural rules, family budgets, personalized nutrition, and real-time pricing.

\section{Proposed Framework}
\label{sec:framework}

\begin{figure}[t]
	\centering
	\includegraphics[width=1\linewidth]{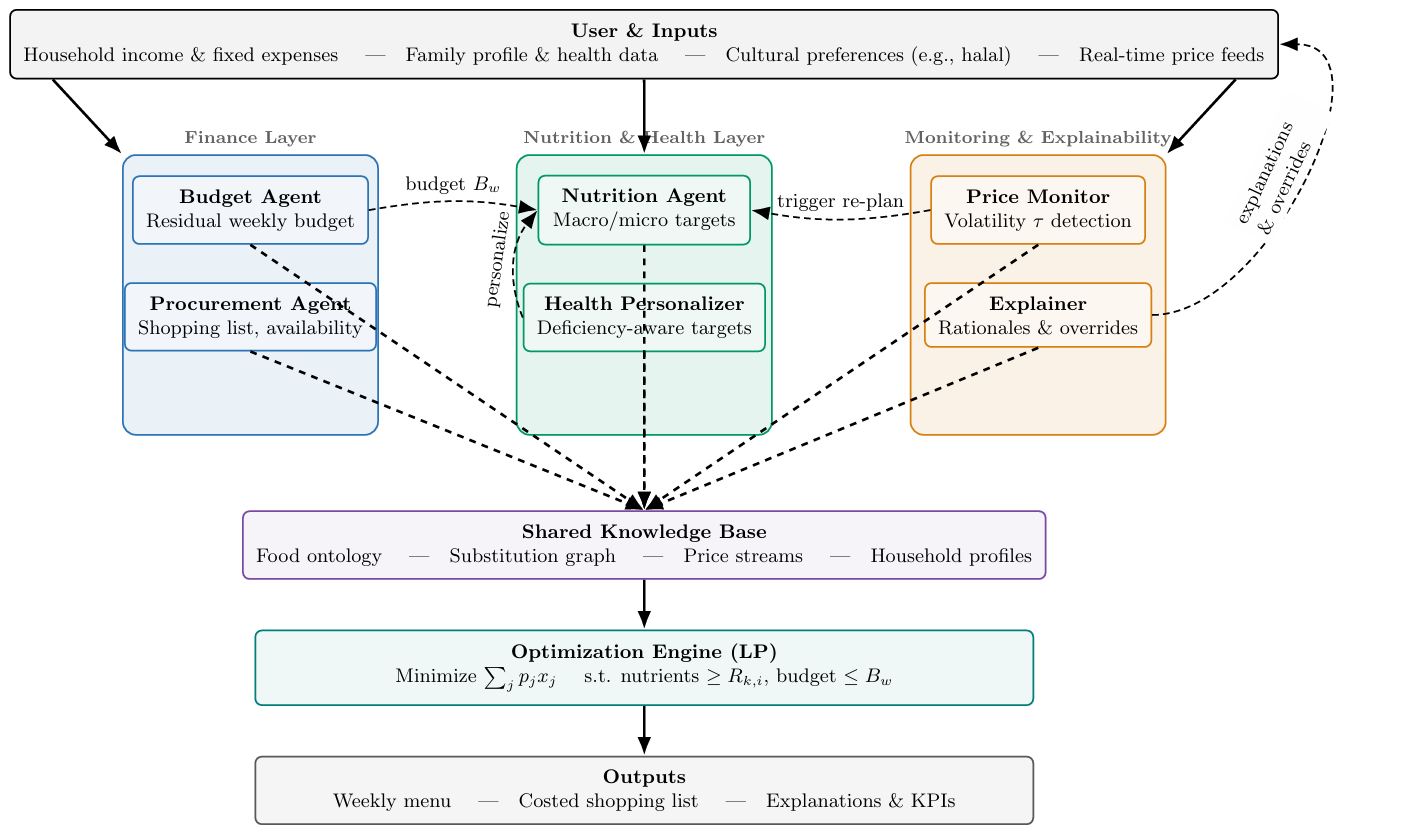}
	\caption{Architecture of the agentic AI framework integrating budgeting, nutrition, cultural preferences, and price awareness}
	\label{fig:fig1}
\end{figure}

The system is an agentic-AI system that jointly optimizes food budgeting and nutritional adequacy in households and responds to supermarket prices, cultural restrictions, and health limitations. A multiagent system will be coordinated into a budgeting, nutrition, cultural knowledge, substitution strategies, and real-time price monitoring system to generate cost-effective weekly meal plans (Figure~\ref{fig:fig1}).

\subsection{System Requirements and Assumptions}

The household income and expenses, demographics, health conditions, nutrient references, cultural rules, and real-time supermarket prices are the inputs in (Table~\ref{tab:inputs}). The weekly food budget $B_w$ is calculated using the disposable income and it is overlaid on aggregate household nutrient needs.

\begin{table}[h!]
	\centering
	\caption{Core input requirements and data sources}
	\label{tab:inputs}
	\begin{tabular}{|p{3cm}|p{5cm}|}
		\hline
		\textbf{Category} & \textbf{Examples / Source} \\ \hline
		Income \& Expenses & Salary, rent, utilities, school fees \\ \hline
		Family Profile & Age, gender, activity level, household size \\ \hline
		Health Constraints & Deficiencies, allergies, sodium limits, diabetic diets \\ \hline
		Price Feeds & Supermarket APIs, real-time scraping \\ \hline
		Nutrition References & WHO, FAO, USDA dietary guidelines \\ \hline
		Cultural Preferences & Halal rules, local cuisine, seasonal foods \\ \hline
	\end{tabular}
\end{table}

\subsection{Multi-Agent System Architecture}

\begin{figure}[t]
	\centering
	\includegraphics[width=1.0\linewidth]{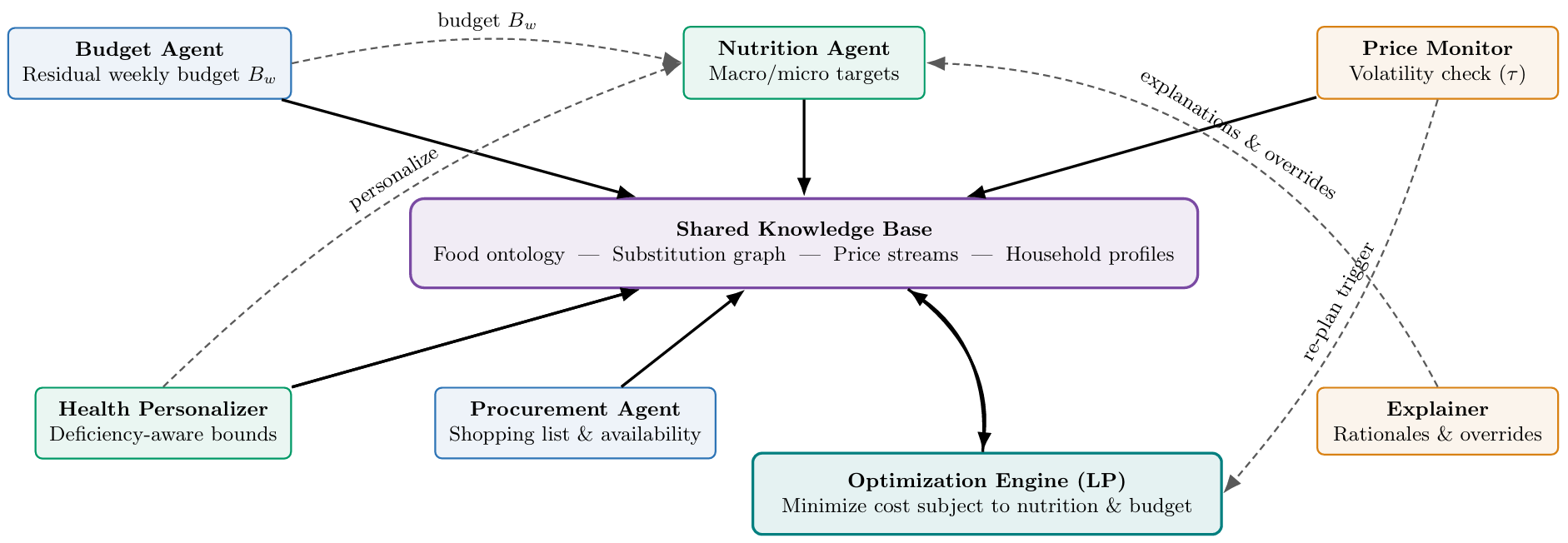}
	\caption{Multi-agent workflow illustrating inter-agent collaboration through the shared knowledge base}
	\label{fig:fig2}
\end{figure}

Agents coordinate via a shared knowledge base (Figure~\ref{fig:fig2}):

\begin{itemize}
	\item \textbf{Budget Agent}: Computes $B_w$.
	\item \textbf{Price Monitor}: Detects price shocks and triggers re-planning.
	\item \textbf{Nutrition Agent}: Ensures macro- and micronutrient adequacy.
	\item \textbf{Health Personalizer}: Adjusts nutrient targets for medical needs.
	\item \textbf{Cultural \& Preference Agent}: Enforces dietary rules.
	\item \textbf{Substitution Agent}: Suggests alternatives via a substitution graph.
	\item \textbf{Procurement Agent}: Converts plans to shopping lists.
	\item \textbf{Explainer Agent}: Provides transparent reasoning.
\end{itemize}

\subsection{Data and Optimization Model}

Agents share a knowledge base with food ontology, substitutions, household profiles, and prices. Meal planning is formulated as a linear program (LP):

\paragraph*{Indices and Variables}
\begin{itemize}
	\item $j \in J$: food items
	\item $n \in N$: nutrients
	\item $m \in H$: household members
	\item $x_j \ge 0$: quantity of food item $j$
\end{itemize}

\paragraph*{Parameters}
\begin{itemize}
	\item $p_j$: item price
	\item $c_{j,n}$: nutrient content
	\item $r_{m,n}$: recommended intake
	\item $R_n = \sum_{m \in H} r_{m,n}$: household requirement
	\item $B_w$: weekly budget
\end{itemize}

\paragraph*{LP Formulation}
\begin{align}
	Z &= \sum_{j \in J} p_j x_j \tag{1} \\[1mm]
	\sum_{j \in J} c_{j,n} x_j &\ge R_n, \quad \forall n \in N \tag{2} \\[1mm]
	\sum_{j \in J} p_j x_j &\le B_w \tag{3} \\[1mm]
	x_j &\ge 0, \quad \forall j \in J \tag{4}
\end{align}

\subsection{Price-Aware Adaptation}

Price shocks are detected if
\[
\frac{|p_{j,t} - p_{j,t-1}|}{p_{j,t-1}} > \tau, \quad \tau \approx 0.10
\]
triggering substitutions and LP re-solving for cost-efficient meal continuity.

\subsection{Health Personalization}

The Health Personalizer calculates $r_{m,n}$ in real time, modifying $R_n$ and constraints, to achieve personalized, adaptive plans.

\subsection{System Workflow}

\begin{figure}[t]
	\centering
	\includegraphics[width=1.0\linewidth]{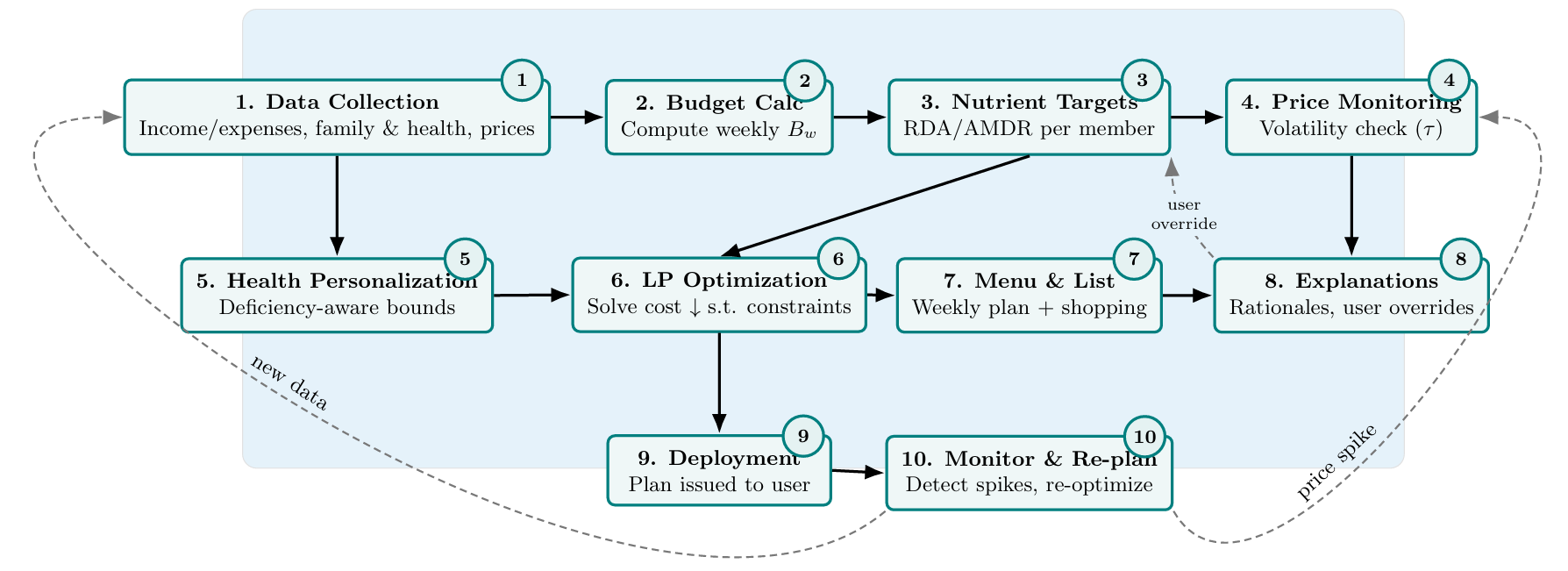}
	\caption{End-to-end workflow from data collection to adaptive re-planning}
	\label{fig:fig3}
\end{figure}

Figure~\ref{fig:fig3} shows the sequence:

\begin{enumerate}
	\item \textbf{Data ingestion}: Populate knowledge base.
	\item \textbf{Budget computation}: Estimate $B_w$.
	\item \textbf{Nutrient profiling}: Compute $R_n$.
	\item \textbf{LP optimization}: Generate baseline plan.
	\item \textbf{Monitoring}: Detect price deviations.
	\item \textbf{Substitution \& re-optimization}: Update LP with alternatives.
	\item \textbf{Plan generation}: Shopping list output.
	\item \textbf{Explanation}: Justification for decisions.
\end{enumerate}

\section{Implementation}
\label{sec:implementation}

Its implementation focuses on modularity, realtime responsiveness, privacy and transparency. The system has subsystems, the LP optimizer, multi-agent orchestration, data management, and price interfaces, which are independent and interact via API, allow scaling, fault isolation and efficient updates.
Table~\ref{tab:implementation} summarizes the main system components and the technologies used for each.

\begin{table}[t]
	\centering
	\caption{Implementation summary of major system components}
	\label{tab:implementation}
	\begin{tabular}{|p{3.0cm}|p{5.0cm}|}
		\hline
		\textbf{Category} & \textbf{Technology / Source} \\ \hline
		Optimization Engine & Python (PuLP, OR-Tools) \\ \hline
		Agent Orchestration & Multi-agent LLM coordination with rule-based triggers \\ \hline
		Database Layer & Household profiles, nutritional tables, health metadata \\ \hline
		Frontend Interface & Lightweight API-driven web UI \\ \hline
		Price Integration & Real-time supermarket APIs and verified scrapers \\ \hline
		Cloud Infrastructure & Containerized deployment with caching and autoscaling \\ \hline
	\end{tabular}
\end{table}

\subsection{System Workflow and Integration}

The LP engine provides real-time nutritional, cultural and budgetary constraints resulting in interpretable solutions. The orchestration layer deals with the communication between agents through event-based stimuli and common knowledge. Structured data are stored in relational databases; high frequency signals are stored in memory. The UI inputs are taken and weekly plans visualized; all the decision-making is made by autonomous agents. The concept of privacy-first is used.

\subsection{Data Sources}

Traceability and repeatability are guaranteed by three distinct data types:

\paragraph*{1) Real Supermarket Data}  
Price feeds by Panda, Carrefour and Lulu are generated using authenticated API and vetted scrapers containing price history of individual items, availability indicators and category attributes. Updated daily.

\paragraph*{2) Authoritative Nutritional References}  
Nutrient values of USDA, FAO, and WHO will be harmonized to GCC dietary standards. All items are mapped using a unified ontology.

\paragraph*{3) Synthetic Household Profiles}  
Households having incomes between 5,000-15,000 SAR, 2-6 individuals, dietary habits (halal, vegetarian, lowsodium), and health conditions influencing nutrient requirement are created by means of stratified sampling using World Bank and CEIC distributions ~\cite{ceic2025globaldata}. Simulation of market volatility is on price shocks of $\pm 10\%$–$\pm 30\%$. Fixed randomly selected seeds will yield reproducibility.

\subsection{Evaluation Protocol}

Two complementary assessments are done, synthetic simulations and a real Saudi household case study.

\paragraph*{1) Synthetic Experiments}  
100 synthetic households are experimented in 12 conditions of price-shocks, repeated 10 times. Metrics include:
\begin{itemize}
	\item \textbf{Cost efficiency}: savings versus static baseline,
	\item \textbf{Nutritional adequacy}: proportion of $R_n$ attained,
	\item \textbf{Adaptivity}: re-planning success rate.
\end{itemize}
Variance and confidence intervals are reported.

\paragraph*{2) Real Household Case Study}  
A four-week comparison of an adaptive vs. static plans of reducing costs, covering nutrients, and practical feasibility is done through a Saudi household. Anonymized non-sensitive data is kept only.

This design facilitates the controlled quantitative analysis and validation in real-life.

\section{Evaluation and Performance Metrics}
\label{sec:evaluation}
The three cases used are (i) synthetic simulations, (ii) ablation experiments and (iii) a real Saudi home case study. Measures were cost efficiency, nutritional adequacy and price shock adaptability. Synthetic experiments were done on 200 households (mean $\pm$ SD). The case study was four weeks long; the experiment involved 312 food items.

\subsection{Cost Efficiency}

Cost efficiency was evaluated with respect to fixed weekly menus, fixed optimization, and manual planning. In 30 trials, the agentic system led to a consistent decrease in weekly costs but high nutrient adequacy. Table~\ref{tab:cost} shows savings of 13--18\% relative to fixed menus.

\begin{table}[t]
	\centering
	\caption{Weekly Food Cost Comparison (SAR, 4-Person Household; mean $\pm$ SD)}
	\label{tab:cost}
	\begin{tabular}{|p{2.7cm}|p{1.8cm}|p{1.2cm}|p{2.3cm}|}
		\hline
		\textbf{Method} & \textbf{Mean Cost} & \textbf{Savings} & \textbf{Nutrition Adequacy (\%)} \\ \hline
		Fixed Menu & 480 $\pm$ 15 & --- & 85 \\ \hline
		Static Optimization & 440 $\pm$ 12 & 8\% & 92 \\ \hline
		Manual Planning & 455 $\pm$ 18 & 5\% & 88 \\ \hline
		Agentic AI & 415 $\pm$ 14 & 13--18\% & 97 \\ \hline
	\end{tabular}
\end{table}

\subsection{Nutritional Adequacy}

\begin{figure}[t]
	\centering
	\includegraphics[width=1.0\linewidth]{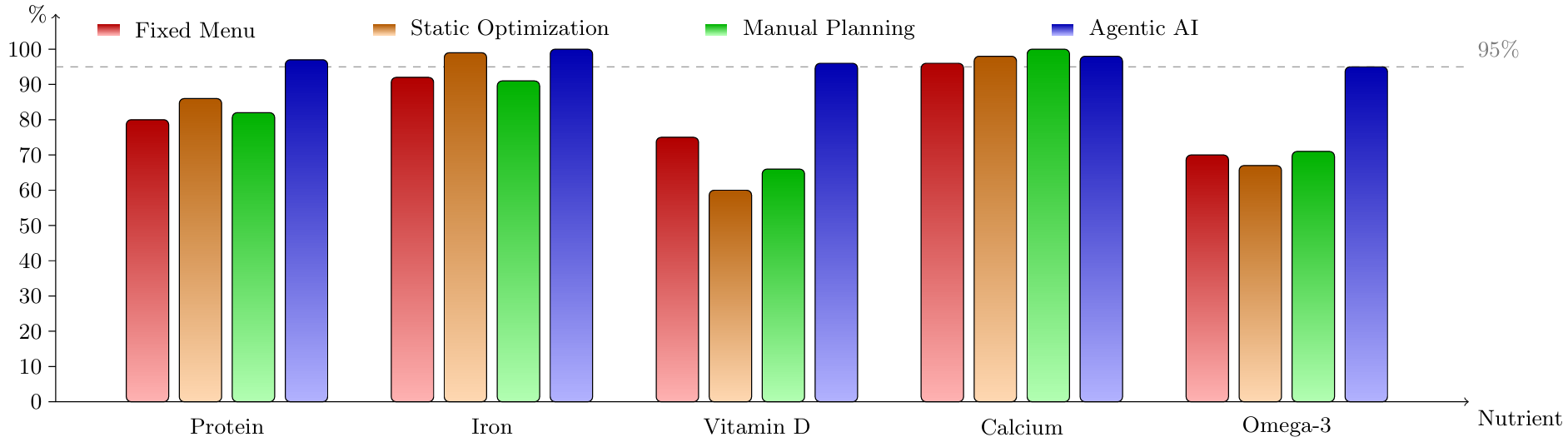}
	\caption{Nutritional adequacy across planning methods (mean over 30 runs).}
	\label{fig:fig5}
\end{figure}

Sufficiency of Protein, Vitamin D, Iron and Calcium was compared with RDAs. The agentic AI attained $\ge$95\% in terms of nutrients, compared to static and manual plans, especially Iron and Vitamin~D (Figure~\ref{fig:fig5}).

\subsection{Adaptivity Under Price Shocks}

\begin{figure}[t]
	\centering
	\includegraphics[width=1.0\linewidth]{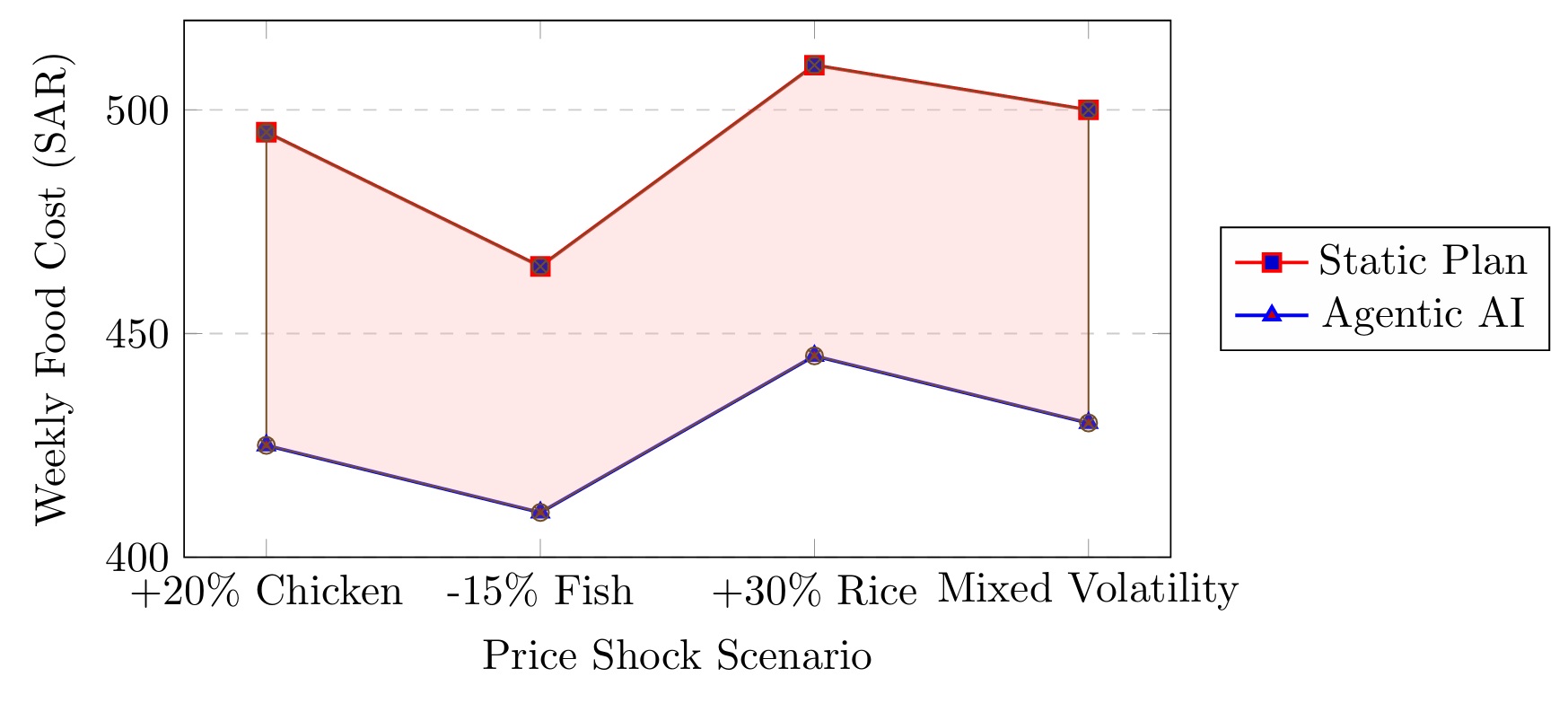}
	\caption{System adaptability under price shocks (SD over 30 runs).}
	\label{fig:fig6}
\end{figure}

Major food were subjected to price shocks of $\pm10\%$, $\pm20\%$, $\pm30\%$. Re-optimization was initiated at $\tau=0.1$. Table~\ref{tab:priceshock} and Figure~\ref{fig:fig6} indicate that the plans of static surpassed the budgets, whereas agentic AI were affordable and sufficient.

\begin{table}[t]
	\centering
	\caption{Adaptivity to Price Shocks (4-Person Household; mean cost over 30 runs)}
	\label{tab:priceshock}
	\begin{tabular}{|p{2.3cm}|p{1.5cm}|p{1.5cm}|p{2.0cm}|}
		\hline
		\textbf{Scenario} & \textbf{Static Plan} & \textbf{Agentic AI} & \textbf{Adequacy Maintained?} \\ \hline
		Chicken +20\% & 495 & 425 & Yes \\ \hline
		Fish --15\% & 465 & 410 & Yes \\ \hline
		Rice +30\% & 510 & 445 & Yes \\ \hline
		Mixed $\pm20\%$ & 500 & 430 & Yes \\ \hline
	\end{tabular}
\end{table}

\subsection{Ablation Studies}

Eliminating the Price Monitor would increase the weekly costs by 9 percent, disabling Health Personalizer dropped Vitamin D adequacy to 70 percent, and eliminating the Preference Agent led to duplicated menus and decreased user satisfaction (3.2 vs. 4.4). Each of the modules is necessary to the performance.

\subsection{Case Study: Saudi Household}

A four-week evaluation with a Saudi household (10{,}000~SAR/month) showed:
\begin{itemize}
	\item \textbf{Average grocery cost:} 1{,}660~SAR (17\% reduction),
	\item \textbf{Nutritional adequacy:} $\geq$95\% including Vitamin~D,
	\item \textbf{Price shock handling:} 20\% chicken increase mitigated via lentil/sardine substitution,
	\item \textbf{User feedback:} cost transparency 4.5/5, cultural relevance 4.2/5.
\end{itemize}

\begin{figure}[t]
	\centering
	\includegraphics[width=0.9\linewidth,height=5cm,keepaspectratio]{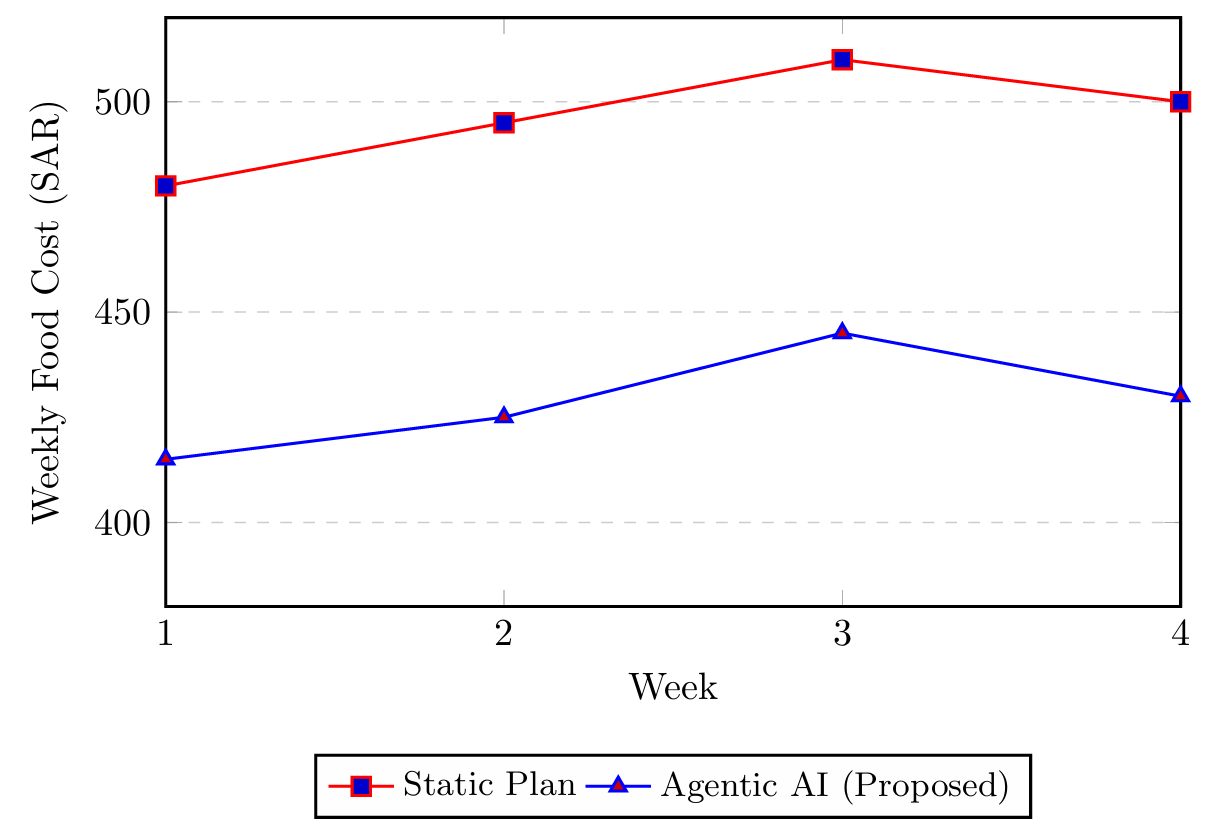}
	\caption{Weekly adaptation of meal plans during the four-week Saudi household case study.}
	\label{fig:fig7}
\end{figure}

Figure~\ref{fig:fig7} illustrates weekly plan adaptations under market changes.

\section{Discussion and Conclusion}
\label{sec:conclusion}

Synthetic simulations, ablation studies, and a four-week Saudi household case study show that the agentic AI framework delivers high nutritional adequacy, cost efficiency, and culturally appropriate meal plans. The actual implementation of the program resulted in a four-member family having $\geq$95\% nutrient coverage, adherence to halal and Ramadan diets, and a 17\% decrease in grocery expenditure.

Robustness is confirmed by experiments with 200 homes over 30 runs: weekly spending decreased by 12–18\% compared to fixed and static baselines, and the system remained stable against $\pm30\%$ price shocks, automatically re-optimizing while maintaining dietary sufficiency.

\subsection*{Limitations and Ethics}

\textbf{Data.} Price feeds that are semi-real-time and comprehensive nutrient tables are required; data that is sparse can be a performance problem. Imputation and region-specific datasets should be included in the future work.

\textbf{Privacy.} Household information is sensitive, and on-device calculation, differential protection, encryption, and retention under the user control reduce risks.

\textbf{Bias.} Recommendations can be skewed with databases or assumptions. They need auditing, constraints of fairness and culturally adaptive food ontologies.


%

\subsection*{Conclusion and Future Work}

Multi-agent system will be used to harmonize budget modeling, nutritional optimization, and price-sensitive adaptation to provide affordable nutritional plans. The next steps involve the large scale deployment, long term compliance studies, as well as connection with loyalty programs, subsidies and school meals, with ethical governance that provides transparency, accountability, and user confidence.

%
%

	\bibliography{references}
	\bibliographystyle{IEEEtran}
	
\end{document}